\documentclass[11pt,a4paper]{article}

\usepackage[final]{acl}
\usepackage{times}
\usepackage{latexsym}

\usepackage{amsmath}
\usepackage{amsfonts}
\usepackage[linesnumbered,ruled,vlined]{algorithm2e}
\usepackage{graphicx} 
\usepackage{booktabs}
\usepackage{array}
\usepackage{enumitem}
\usepackage{amsthm}
\usepackage{algpseudocode}
\usepackage[switch]{lineno}
\usepackage[utf8]{inputenc}
\usepackage{eqparbox}
\usepackage[nopar]{lipsum}
\usepackage{multirow}
\usepackage{makecell}
\usepackage{xcolor}
\usepackage{hhline} 
\usepackage{microtype}

\usepackage{relsize}

\usepackage{booktabs,multirow,array}
\newcolumntype{N}{@{}m{0pt}@{}}

\usepackage[many]{tcolorbox}
\newtcolorbox{fancyquotes}{%
    enhanced jigsaw, 
    breakable,      
    frame hidden,   
    left=0.5cm,       
    right=0.1cm,      
    overlay={%
        \node [scale=8,
            text=black,
            inner sep=0pt,] at ([xshift=-1cm,yshift=-1cm]frame.north west){}; 
        \node [scale=8,
            text=black,
            inner sep=0pt,] at ([xshift=1cm]frame.south east){};  
            },
                parbox=false,
}

\usepackage{mathtools, nccmath}
\usepackage{scrextend}
\deffootnote[.25in]{.25in}{.15in}{\makebox[.25in][r]{\thefootnotemark .\hspace{.15in}}}

\makeatletter

\newtheorem*{proof*}{Proof}

\usepackage{listings}
\usepackage{color}
\definecolor{codegreen}{rgb}{0.3,0.5,0.0}
\def\@fnsymbol#1{\ensuremath{\ifcase#1\or \dagger\or *\or \ddagger\or
   \mathsection\or \mathparagraph\or \|\or **\or \dagger\dagger
   \or \ddagger\ddagger \else\@ctrerr\fi}}

\newcolumntype{C}[1]{>{\centering\let\newline\\\arraybackslash\hspace{0pt}}m{#1}}
\newcommand\ChangeRT[1]{\noalign{\hrule height #1}}

\title{RetroMAE v2: Duplex Masked Auto-Encoder For Pre-Training Retrieval-Oriented Language Models} 

\author{Shitao Xiao$^1$, Zheng Liu$^2$ \\
  1: Beijing University of Posts and Telecommunications, Beijing, China \\ 
  2: Huawei Technologies Ltd. Co., Shenzhen, China \\
  \texttt{stxiao@bupt.edu.cn}, 
  \texttt{zhengliu1026@gmail.com}
}

\begin{document}
\maketitle 


\begin{abstract}
To better support retrieval applications such as web search and question answering, growing effort is made to develop retrieval-oriented language models \cite{gao2021condenser,wang2021tsdae,liu2022retromae}. 
Most of the existing works focus on improving the semantic representation capability for the contextualized embedding of [CLS] token. However, recent study shows that the ordinary tokens besides [CLS] may provide extra information, which helps to produce a better representation effect \cite{lin2022aggretriever}. As such, it's necessary to extend the current methods where all contextualized embeddings can be jointly pre-trained for the retrieval tasks. 


With this motivation, we propose a new pre-training method: duplex masked auto-encoder, \textit{a.k.a.} DupMAE, which targets on improving the semantic representation capacity for the contextualized embeddings of both [CLS] and ordinary tokens. It introduces two decoding tasks: one is to reconstruct the original input sentence based on the [CLS] embedding, the other one is to minimize the bag-of-words loss (BoW) about the input sentence based on the entire ordinary tokens' embeddings. The two decoding losses are added up to train a unified encoding model. The embeddings from [CLS] and ordinary tokens, after dimension reduction and aggregation, are concatenated as one unified semantic representation for the input. DupMAE is simple but empirically competitive: with a small decoding cost, it substantially contributes to the model's representation capability and transferability, where remarkable improvements are achieved on MS MARCO and BEIR benchmarks.

\end{abstract} 

\section{Introduction}


Deep semantic retrieval is important to many real-world scenarios, such as web search, question answering and conversational system \cite{huang2013learning,karpukhin2020dense,komeili2021internet,izacard2022few}. In recent years, pre-trained language models, e.g., BERT \cite{Devlin2019BERT}, RoBERTa \cite{Liu2019Roberta}, T5 \cite{raffel2019exploring}, are widely adopted as the retrievers' backbone networks. The generic pre-trained language models are not directly applicable to retrieval tasks. Thus, it calls for complex fine-tuning strategies, such as advanced negative sampling \cite{xiong2020approximate,qu2020rocketqa}, knowledge distillation \cite{hofstatter2021efficiently,lu2022ernie} and joint learning \cite{ren2021rocketqav2,zhang2021adversarial}. To reduce this effort and bring in better retrieval quality, there are growing interests in developing retrieval-oriented language models. One common practice is to leverage self-contrastive learning \cite{chang2020pre,guu2020realm}, where the language models are learned to discriminate heuristically acquired positive and negative samples in the embedding space. Later on, auto-encoding is found to be more effective \cite{wang2021tsdae,lu2021less}, where the language models are learned to reconstruct the input based on the generated embeddings. The recent work RetroMAE \cite{liu2022retromae} extends the previous auto-encoding methods by introducing the enhanced encoding and decoding mechanisms, which leads to remarkable improvements on general retrieval benchmarks.

The existing retrieval-oriented pre-trained models mainly rely on the contextualized embedding from the special token, i.e., [CLS], to represent the semantic about input \cite{gao2021condenser,lu2021less,liu2022retromae,wang2022simlm}. However, recent study finds that other ordinary tokens may provide extra information and help to generate better semantic representations \cite{lin2022aggretriever}. Such a statement is consistent with previous research \cite{luan2021sparse,santhanam2021colbertv2}, as multi-vector or token-granularity representations may give higher discriminative power than those based on one single vector. As a result, it is necessary to extend the current works, such that the representation capability can be jointly pre-trained for both [CLS] and ordinary tokens. 

\begin{figure}[t]
\centering
\includegraphics[width=1.0\linewidth]{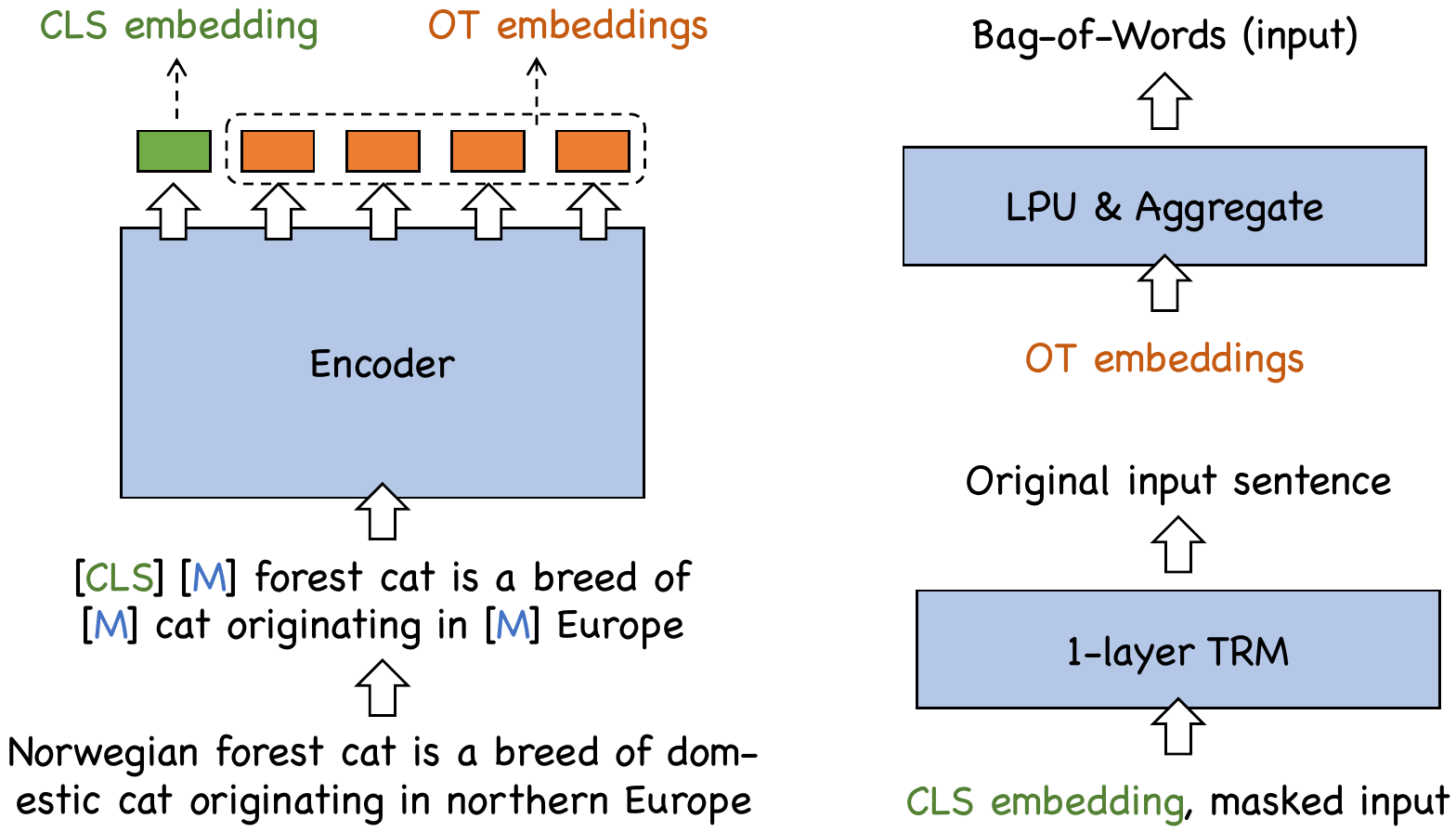}
\caption{{DupMAE}. Encoder: the sentence is masked and encoded as the contextualized embeddings for [CLS] and ordinary tokens (OT). Decoder: the CLS embedding is joined with the masked input, where the original input is recovered by a one-layer transformer; OT embeddings are transformed into vocabulary space via LPU and aggregated to preserve the BoW feature.} 
\label{fig:1}
\end{figure}

With this motivation, we propose a new retrieval-oriented pre-training method Duplex Masked Auto-Encoder, \textit{a.k.a.} \textbf{DupMAE}, shown as Figure \ref{fig:1}.  

$\bullet$ \textbf{Workflow}. DupMAE contains a unified encoder, which produces the contextualized embeddings for both [CLS] and ordinary tokens for the masked input. It introduces two decoding modules, which work together to enhance the semantic representation capacity for both types of contextualized embeddings. Particularly, we leverage the decoder from RetroMAE \cite{liu2022retromae}, where the [CLS] embedding, joined with the masked input, is used to recover the original sentence via an one-layer transformer. Meanwhile, the contextualized embeddings from ordinary tokens are transformed into the vocabulary space (i.e, $|V|$-dim vectors) via a linear projection unit (i.e., a $d \times |V|$ matrix). The transformation results from all ordinary tokens are aggregated by max-pooling, where the BoW (Bog-of-Words) feature about the input is preserved.  

$\bullet$ \textbf{Merits}. The above workflow is highlighted by its {extremely} {simplified} decoders: a single-layer transformer to recover the original sentence, and a linear projection unit to preserve the BoW feature. It brings two merits as a consequence. Firstly, the pre-training is made \texttt{Low-cost} given that the decoding operations are simple. Secondly and more importantly, the pre-training task is made highly \texttt{Demanding}: as the decoders are extremely weak due to simplicity, it will force the encoder to fully preserve the input information so as to produce high-fidelity recovery of the input. 


$\bullet$ \textbf{Representation}. The contextualized embeddings from [CLS] and ordinary tokens are aggregated in a simple way for the final representation. The [CLS] embedding is reduced to a lower dimension by linear projection. The ordinary tokens' embeddings, after transformed into the vocabulary space and aggregated by max-pooling, are sparsified by picking the top-N entries. The two results are concatenated as one vector. With properly configured linear projection and sparsification, it will preserve a similar distance computation cost and memory footprint as the original embedding.  

Our proposed method is simple but empirically competitive. We use common pre-training data for DupMAE (Wikipedia, BookCorpus, MS MARCO), where a BERT-base scale encoder is produced. 
Empirical study shows that the semantic representation capacities can be substantially improved for both [CLS] and ordinary tokens' embeddings, as remarkable performances are generated throughout different scenarios, e.g., a MRR@10 of \textbf{42.6} on \textbf{MS MARCO} (supervised retrieval), and an average NDCG@10 of \textbf{47.5} on all 18 datasets of \textbf{BEIR} (zero-shot retrieval). Our models and code will be available at \cite{retromae-code} for public access.  


\begin{figure*}[t]
\centering
\includegraphics[width=1.0\textwidth]{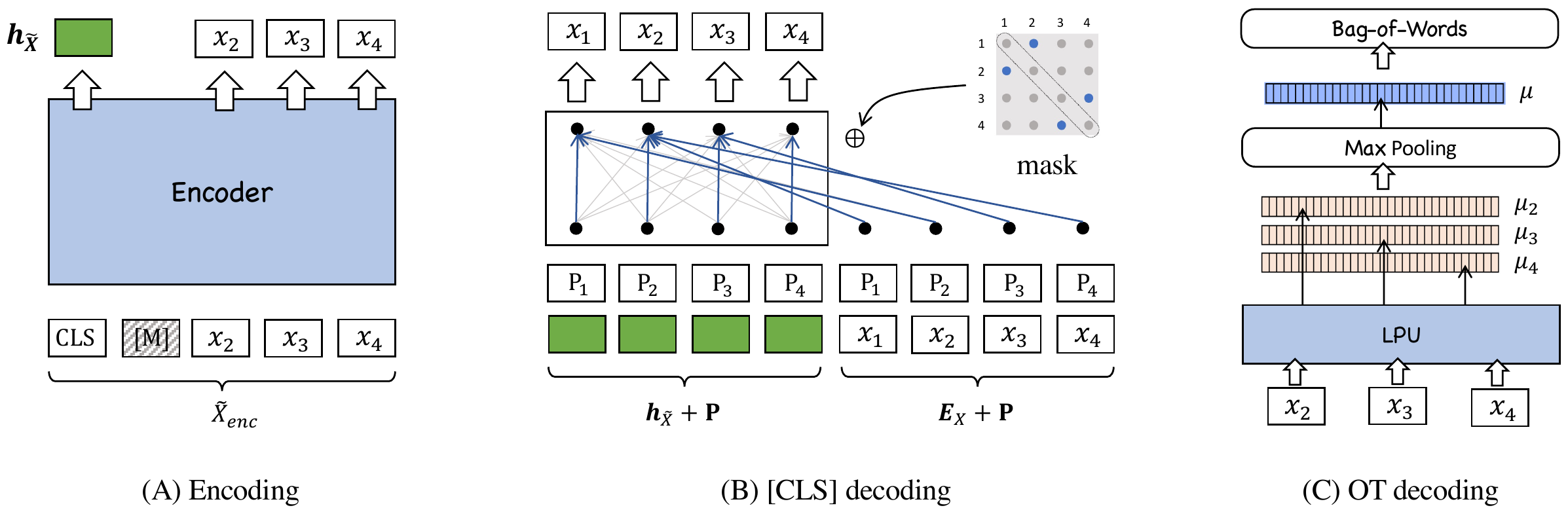}
\vspace{-15pt}
\caption{Framework of DupMAE. The unified encoder generates the contextualized embeddings for the [CLS] and ordinary tokens (OT). The [CLS] decoding reconstructs the original sentence based on an one-layer transformer; the OT decoding preserves the BoW feature of the input based on LPU and max-pooling.}   
\vspace{-10pt}
\label{fig:2}
\end{figure*} 


\section{Related Works}

Deep semantic retrieval is a critical component to many important applications, such as web search, question answering, advertising and recommender systems \cite{karpukhin2020dense,zhang2022uni,xiao2022training}. It maps the query and document into embeddings within latent space, where the semantic relevance can be measured by the embedding similarity. In recent years, the pre-trained language models have been widely applied to deep semantic retrieval such that discriminative representations can be generated for the queries and documents. Despite the preliminary progress achieved by early pre-trained models, like BERT \cite{Devlin2019BERT}, it is noticed that the more advanced models bring little extra benefit to the retrieval quality, and it's believed that conventional pre-training algorithms are not compatible with the purpose of deep semantic retrieval \cite{gao2021condenser,lu2021less,wang2022simlm}. 

To mitigate the above problem, people become increasingly interested in developing retrieval oriented pre-trained models. For example, it is proposed to leverage self-contrastive learning (SCL) where the language models are pre-trained to discriminate positive samples generated by data augmentation and in-batch negative samples \cite{chang2020pre,guu2020realm,izacard2021towards}. The SCL based algorithms are limited due to the quality of data augmentation, and the requirement of huge amounts of negative samples. Later on, the auto-encoding based pre-training algorithms receive growing interests: the input sentences are encoded into embeddings and reconstructed back to the original sentences \cite{lu2021less,wang2021tsdae}. The recently proposed methods, such as SimLM \cite{wang2022simlm} and RetroMAE \cite{liu2022retromae}, extend the previous auto-encoding framework by upgrading the encoding and decoding mechanisms, which substantially improves the quality of deep semantic retrieval. 

The existing retrieval-oriented pre-training methods target on improving the semantic representation capacity for the contextualized embedding from the [CLS] token. However, it is noticed that the ordinary tokens may provide additional information besides [CLS], especially when dealing with long and semantic-rich documents \cite{luan2021sparse,humeau2019poly,lin2022aggretriever}. As a result, it is necessary to extend the current works, where the semantic representation capability can be enhanced for both types of contextualized embeddings.

\section{Methodology}
In this section, we'll start with an overview of DupMAE. Then, we'll present the preliminaries about RetroMAE, which is the foundation of DupMAE. 


The framework of DupMAE is shown as Figure \ref{fig:2}. There is an unified encoder (A), where the masked sentence is encoded into the contextualized embeddings. There are two decoding components. One is for [CLS] decoding (B): it is based on a single-layer transformer, which reconstructs the original sentence based on the [CLS] embedding. The other one is for OT decoding: it is based on a linear projection unit LPU (C), which transforms the ordinary token embeddings into the vocabulary space and aggregates them via max-pooling to preserve the BoW feature of the input. The embeddings from CLS and ordinary tokens are jointly learned based on the two decoders. With reduced dimensions, the whole embeddings are aggregated to represent the input, such that the memory footprint will be similar as using one single embedding. 


\subsection{Preliminaries of RetroMAE}
The encoding and decoding workflow of RetroMAE \cite{liu2022retromae} are shown as Figure (A) and (B). 
An input sentence $X$ is sampled and masked as $\tilde{X}_{enc}$ by randomly replacing some of its tokens with the special token [M]. A small masking ratio is applied for encoding (15$\sim$30\%), where the major input information can be preserved. The encoder $\Phi^{enc}(\cdot)$ is used to transform the masked sentence into the sentence embedding $\mathbf{h}_{\tilde{X}}$: 
\begin{equation}
    \mathbf{h}_{\tilde{X}} \leftarrow \Phi_{enc}(\tilde{X}_{enc}). 
\end{equation}
In order to capture the in-depth semantics about the sentence, a full-scale BERT-base encoder is used to generate to the sentence embedding. The contextualized embedding from [CLS] token is used as the sentence embedding. The masked tokens are predicted following the typical form of masked language modeling \cite{Devlin2019BERT}, which gives rise to the encoder's loss: $\mathcal{L}_{mlm}$. 

As for decoding, the input sentence $X$ is masked again as $\tilde{X}_{dec}$. A much more aggressive masking ratio is applied, where the majority of the tokens (50$\sim$90\%) will be masked. The masked input is joined with the sentence embedding, based on which the original sentence is reconstructed. RetroMAE is highlighted for its enhanced decoding, where the two-stream self-attention and the position-specific attention mask are utilized. Particularly, it generates two input streams: $\mathbf{H}_1$ (query) and $\mathbf{H}_2$ (context), for decoding:
\begin{equation}
\begin{gathered}
\label{eq:4}
\mathbf{H}_1 \leftarrow [\mathbf{h}_{\tilde{X}} + \mathbf{p}_0,...,
\mathbf{h}_{\tilde{X}} + \mathbf{p}_N], \\
\mathbf{H}_2 \leftarrow  
[\mathbf{h}_{\tilde{X}}, \mathbf{e}_{x_1}+\mathbf{p}_1, ... , \mathbf{e}_{x_N}+\mathbf{p}_N]. 
\end{gathered}
\end{equation}
where $\mathbf{h}_{\tilde{X}}$ is the sentence embedding, $\mathbf{e}_{x_i}$ is the token embedding, $\mathbf{p}_i$ is the position embedding.
We introduce the position-specific attention mask $\mathbf{M} \in \mathbb{R}^{L \times L}$, where self-attention is performed as:
\begin{equation}\label{eq:5}
\begin{gathered}
    \mathbf{Q} = \mathbf{H}_1\mathbf{W}^Q, \mathbf{K} = \mathbf{H}_2\mathbf{W}^K, \mathbf{V} = \mathbf{H}_2\mathbf{W}^V; \\
    \mathbf{M}_{ij} = 
    \begin{cases}
    0, ~~~~~~\text{can be attended}, \\
    -\infty, ~\text{masked}; 
    \end{cases} \\
    \mathbf{A} = \mathrm{softmax}(\frac{\mathbf{Q}^T\mathbf{K}}{\sqrt{d}} + \mathbf{M
    })\mathbf{V}. 
\end{gathered}
\end{equation}
The output $\mathbf{A}$, together with $\mathbf{H}_{1}$ (from the residual connection) are used to reconstruct the original input.
Finally, the following objective is optimized: 
\begin{equation}\label{eq:6}
 \mathcal{L}_{dec} = \sum\nolimits_{x_i \in X} \mathrm{CE}(x_i|\mathbf{A}, \mathbf{H}_{1}).
\end{equation} 
As the decoder only contains one transformer layer, each token $x_i$ is reconstructed based on the unique context which are visible to the $i$-th row of $\mathbf{M}$. In this place, $\mathbf{M}$ is generated by the following rules:  
\begin{equation}\label{eq:7}
\begin{gathered}
\mathbf{M}_{ij} = 
    \begin{cases}
    0,   ~~ x_j \in s(X_{\neq i}), ~\text{or}~ j_{{|i\neq0}}=0 \\
    -\infty,  ~~ \text{otherwise},
    \end{cases} 
\end{gathered}
\end{equation}
which means the sampled tokens $s(X_{\neq i})$, and the first positions will be visible ($0$). Meanwhile, the diagonal elements will always be masked ($-\infty$), such that each token may not attend to itself.


\subsection{Extension to DupMAE}\label{sec:method-dup}
The spirit of RetroMAE can be summarized by the following statement: the encoder needs to fully preserve the input information by the contextualized embeddings, with which the original input can be effectively reconstructed via extremely simplified decoders. We extend the spirit to DupMAE, where two different decoders are deployed for the contextualized embeddings of [CLS] and OT. For one thing, it leverages the encoding and decoding workflow of RetroMAE for the [CLS] embedding. For another thing, it introduces a new decoding strategy to pre-train the ordinary tokens' (OT) embeddings, as Figure \ref{fig:2} (C). It involves two basic operations. Firstly, the OT embeddings $\mathbf{E}_{\tilde{X}_{enc}}$: $\{\mathbf{e}_{x_1}, ..., \mathbf{e}_{x_N}\}$ (masked tokens excluded) are mapped to the vocabulary space via the linear projection unit: 
\begin{equation}\label{eq:8}
    \boldsymbol{\mu}_{x_i} \leftarrow \mathbf{e}_{x_i}^T\mathbf{W}^O, 
    ~ x_i \in \tilde{X}_{enc},
\end{equation}
where $\mathbf{W}^O \in \mathbb{R}^{d \times |V|}$, $d$: the dimension of embedding, $|V|$: the size of vocabulary. Then, the transformation results of OT embeddings are aggregated via token-wise max-pooling:
\begin{equation}\label{eq:9}
    \boldsymbol{\mu}_{\tilde{X}_{enc}} \leftarrow token.\mathrm{Max}(\{\boldsymbol{\mu}_{x_i}|\tilde{X}_{enc}\}).
\end{equation}
In other words, the maximum values of all tokens in $\{\boldsymbol{\mu}_{x_i}|\tilde{X}_{enc}\}$ will be preserved for each vocabulary. We expect the BoW feature of the input to be preserved by $\boldsymbol{\mu}_{x_i} \in \mathbb{R}^{|V|}$. Therefore, the following cross entropy loss $\mathcal{L}_{BoW}$ is minimized: 
\begin{equation}\label{eq:10}
    \mathrm{min}. - \sum_{x \in set(X)} \log 
    \frac{\exp(\boldsymbol{\mu}_{\tilde{X}_{enc}}[x])}
    {\sum_{x' \in V} \exp(\boldsymbol{\mu}_{\tilde{X}_{enc}}[x'])},
\end{equation}
where $x\in set(X)$ is a unique vocabulary in the input $X$, $V$ indicates the whole vocabulary. The losses from encoder, [CLS] (Eq. \ref{eq:6}) and OT decoding (Eq. \ref{eq:10}) are added up as our objective: 
\begin{equation}\label{eq:11}
    \mathrm{min}. ~ \mathcal{L}_{mlm} + \mathcal{L}_{dec} + \mathcal{L}_{BoW}.
\end{equation}

\textbf{Representation}. A remaining problem of DupMAE is how to generate the semantic representation for the input. Particularly, it's desired to fully leverage the capacities from both [CLS] and OT embeddings, where a stronger representation can be produced. Besides, it has to be compact, where the retrieval process can be memory and temporally efficient. With both considerations, we propose the aggregation strategy. Firstly, the [CLS] embedding $\mathbf{h}_X$ is linearly mapped to a lower dimension: 
\begin{equation}\label{eq:12} 
    \mathbf{\hat{h}}_X \leftarrow \mathbf{h}_X^T \mathbf{W}^{cls}, ~
    \mathbf{W}^{cls} \in \mathbb{R}^{d \times d'}.
\end{equation}
Secondly, we leverage $\boldsymbol{\mu}_{X}$ to aggregate the information from OT embeddings, where we reduce its dimension via sparsification: 
\begin{equation}\label{eq:13}
    \boldsymbol{\hat{\mu}}_{X} 
    \leftarrow
    \{i: \boldsymbol{\mu}_{X}[i] ~| ~ i \in 
    I_X \}.
\end{equation}
Here, $I_X$ stands for the indexes where $\boldsymbol{\mu}_{X}[i] \in \text{top-k}(\boldsymbol{\mu}_{X})$, $k$ is the number of elements to be preserved for $\boldsymbol{\mu}_{X}$. For each document, we combine the dim-reduction results of [CLS] and OT embeddings as its semantic representation: $[\mathbf{\hat{h}}_X; \boldsymbol{\hat{\mu}}_{X}]$. For a give query, we measure its relevance to the document with the following inner-product: 
\begin{equation}\label{eq:14} 
    \langle q,d \rangle = \mathbf{\hat{h}}_q^T \mathbf{\hat{h}}_d + 
    \sum\nolimits_{I_d} \boldsymbol{\mu}_q[i] \boldsymbol{\mu}_d[i].
\end{equation} 
With properly configured dim-reduction, the memory footprint and the cost of relevance computation will be similar with the convention methods.  

\textbf{Fine-Tuning}. After pre-training, we fine-tune the encoder with three steps. Firstly, the contrastive loss is minimized for the in-batch negatives ($\text{IB}$): 
\begin{equation}\label{eq:15}
    \min. - \sum_{q} \log \frac{\exp(\langle q,d^+ \rangle)}
    {\sum_{d \in \{d^+, \text{IB}\} }\exp(\langle q,d \rangle)}. 
\end{equation}
In the 2nd step, we collect the ANN hard negatives ($D^-$) for each query based on the first-stage encoder, and continue to minimize the contrastive loss with both in-batch and hard negatives: 
\begin{equation}\label{eq:16}
    \min. - \sum_{q} \log \frac{\exp(\langle q,d^+ \rangle)}
    {\sum\nolimits_{d \in \{d^+, D^-, \text{IB} \} }\exp(\langle q,d \rangle)}. 
\end{equation} 
In the 3rd step, we use knowledge distillation: a cross-encoder is trained to discriminate the positives ($d^+$) from negatives ($d^-$) for each query; then, the soft labeled cross-entropy is minimized: 
\begin{equation}\label{eq:17}
    \min. - \sum_q \sigma_q^d \log \frac{\exp(\langle q,d^+ \rangle)}
    {\sum_{d \in \{d^+, D^-\} }\exp(\langle q,d \rangle)}
\end{equation} 
where $\sigma_q^d$ is the softmax activation of the cross-encoder's prediction of q and d. 

Fine-tuning with the first two steps is cost-effective, as the encoder is trained purely with low-cost operations. The third step brings much higher cost due to the training and soft-labeling of the cross-encoder. However, it helps to further improve the encoder's retrieval quality. In our experimental studies, comprehensive analysis is made for DupMAE's impact in different stages. 

\begin{table*}[t]
    \centering
    \scriptsize
    \begin{tabular}{p{3.9cm}|C{1.0cm}|C{1.1cm}|C{1.1cm}|C{1.1cm}|C{1.2cm}|C{1.2cm} }
    \ChangeRT{1pt} 
    & &
    \multicolumn{3}{c|}{\textbf{MS MARCO Dev}} & \multicolumn{1}{c|}{\textbf{DL'19}} &
    \multicolumn{1}{c}{\textbf{DL'20}} \\
    \cmidrule(lr){1-1}
    \cmidrule(lr){2-2}
    \cmidrule(lr){3-5}
    \cmidrule(lr){6-6}
    \cmidrule(lr){7-7}
    \textbf{Methods} & \textbf{FT} &
    \textbf{MRR@10} & \textbf{R@50} & \textbf{R@1000} & \textbf{NDCG@10} & \textbf{NDCG@10}  \\
    \hline
    ANCE \cite{xiong2020approximate} & hard & 0.330 & -- & 0.959 & 0.648 & 0.615 \\
    SEED \cite{lu2021less} & hard & 0.339 & -- & 0.961 & -- & -- \\
    ADORE \cite{zhan2021optimizing} & hard & 0.347 & -- & -- & 0.683 & -- \\
    Condenser \cite{gao2021condenser} & hard & 0.366 & -- & 0.974 & 0.698 & -- \\
    coCondenser \cite{gao2021unsupervised} & hard & 0.382 & -- & 0.984 & 0.717 & 0.684 \\
    \hline
    TAS-B \cite{hofstatter2021efficiently} & distill & 0.340 & -- & 0.975 & 0.712 & 0.693 \\
    RocketQAv2 \cite{ren2021rocketqav2} & distill & 0.388 & 0.862 & 0.981 & -- & -- \\
    AR2 \cite{zhang2021adversarial} & distill & 0.395 & 0.878 & 0.986 & -- & -- \\
    AR2+SimANS \cite{zhou2022simans} & distill & 0.409 & 0.887 & 0.987 & -- & -- \\
    SPLADEv2 \cite{formal2021splade} & distill & 0.368 & -- & 0.979 & 0.729 & -- \\
    ColBERTv2 \cite{santhanam2021colbertv2} & distill & 0.397 & 0.868 & 0.984 & -- & -- \\
    ERNIE-Search \cite{lu2022ernie} & distill & 0.401 & 0.877 & 0.982 & -- & -- \\
    SimLM \cite{wang2022simlm} & distill & 0.411 & 0.878 & 0.987 & 0.714 & 0.697 \\
    RetroMAE \cite{liu2022retromae} & distill & 0.416 & 0.885 & 0.988 & 0.681 & 0.706  \\
    \hhline{=|=|=|=|=|=|=}
    DupMAE (stage 2) & hard & 0.4102 & 0.8875 & 0.9874 & 0.7128 & \textbf{0.7095} \\
    DupMAE (stage 3) & distill & \textbf{0.4258} & \textbf{0.8966} & \textbf{0.9893} & \textbf{0.7509} & {0.7083} \\
    \ChangeRT{1pt}
    \end{tabular}
    \vspace{-5pt}
    \caption{Comparisons between DupMAE (from stage 2 and stage 3) and competitive baseline methods in recent years (``hard'': fine-tuned by hard negative samples, ``distill'': fine-tuned by knowledge distillation).} 
    \vspace{-10pt}
    \label{tab:2}
\end{table*}

\section{Experiment}
The empirical studies are conducted to explore the following research questions. 
\begin{itemize}
    \item \textbf{RQ 1.} Whether DupMAE produces better semantic representations, especially compared with its close peer RetroMAE?
    \item \textbf{RQ 2.} Whether DupMAE is able to maintain its advantages throughout different situations?
    \item \textbf{RQ 3.} Whether DupMAE benefits from the joint utilization of [CLS] and OT embeddings, and what's the individual contribution from each embedding?
    \item \textbf{RQ 4.} Whether the pre-training tasks contribute to both [CLS] and OT embeddings?
\end{itemize}

\textbf{Datasets}. The retrieval performances are evaluated under both supervised and zero-shot settings. We choose the passage retrieval task of \textbf{MS MARCO} \cite{nguyen2016ms} for our supervised evaluation. It contains real-world queries from Bing Search. The ground-truth answer to the query needs to be retrieved from 8.8 million candidate passages. We use three sets of queries for evaluation: the Dev set of MS MARCO, and the test sets for TREC Deep Learning track in 2019 and 2020, denoted as DL'19 and DL'20 \cite{craswell2020overview}. We leverage \textbf{BEIR} \cite{thakur2021beir} for our zero-shot evaluation. It contains a total of 18 datasets, which covers diverse types of retrieval tasks, such as question answering, entity retrieval, fact verification, etc. The pre-trained models are fine-tuned with queries from MS MARCO, and evaluated for their out-domain retrieval performances on each of the 18 datasets.


\textbf{Baselines}. We consider two types of baselines for supervised evaluations. The first type introduces models fine-tuned based on \textbf{hard negative samples}, including ANCE \cite{xiong2020approximate}, SEED \cite{lu2021less}, ADORE \cite{zhan2021optimizing}, Condenser \cite{gao2021condenser}, and coCondener \cite{gao2021unsupervised}. The second type contains models fine-tuned based on \textbf{knowledge distillation}, such as TAS-B \cite{hofstatter2021efficiently}, RocketQAv2 \cite{ren2021rocketqav2}, AR2 \cite{zhang2021adversarial}, AR2+SimANS \cite{zhou2022simans}, SPLADEv2 \cite{formal2021splade}, ColBERTv2 \cite{santhanam2021colbertv2}, ERNIE-Search \cite{lu2022ernie}, SimLM \cite{wang2022simlm}, RetroMAE \cite{liu2022retromae}. At the same time, we consider two types of approaches for zero-shot evaluations. The first type is the commonly used sparse retrieval baseline \textbf{BM25}. The second type include dense retrievers based on \textbf{pre-trained language models}, like BERT \cite{Devlin2019BERT} and RetroMAE \cite{liu2022retromae}. Among them, Contriever \cite{izacard2021towards} and GTR-* \cite{ni2021large} are pre-trained with massive amounts of data through contrastive learning, and GTR-XXL is a super large model with 4.8 billion parameters (more than 40$\times$ larger than other methods). 


\textbf{Implementation details}. RetroMAE utilizes bi-directional transformers as its encoding network, with 12 layers, 768 hidden-dim, and a vocabulary of 30522 tokens (same as BERT base). The decoder is a one-layer transformer. The [CLS] embedding and OT embedding are reduced to dim-384 by default; as a result, it will result in the same similarity computation cost as the majority of baselines which uses a dim-768 embedding. We also explore other configurations of dimensions in our experiments. The masking ratio is set to 0.3 for encoder and 0.5 for decoder. We leverage three commonly used corpora for pre-training: Wikipedia, BookCorpus \cite{Devlin2019BERT}, and MS MARCO \cite{nguyen2016ms}. The pre-training and fine-tuning take place on machines with 8$\times$ Nvidia V100 (32GB) GPUs. The models are implemented with PyTorch 1.8 and HuggingFace transformers 4.16. 


\begin{table*}[t]
    \centering
    \scriptsize
    \begin{tabular}{p{1.7cm}|C{0.8cm}|C{0.8cm}|C{0.8cm}|C{1.2cm}|C{1.2cm}|C{1.2cm}|C{1.2cm}|C{1.2cm}|C{1.2cm} }
    \ChangeRT{1pt}
    \textbf{Datasets} & \textbf{BM25} & 
    \textbf{BERT} & \textbf{SEED} & \textbf{Condenser} & \textbf{Contriever} & \textbf{GTR-base} & \textbf{GTR-XXL} & \textbf{RetroMAE} & \textbf{DupMAE} \\
    \hhline{=|=|=|=|=|=|=|=|=|=}
    TREC-COVID & 0.656 & 0.615 & 0.627 & 0.750 & 0.596 & 0.539 & 0.501 & \textbf{0.772} & 0.728 \\
    BioASQ & 0.465 & 0.253 & 0.308 & 0.322 & 0.383 & 0.271 & 0.324 & 0.421 & \textbf{0.508} \\
    NFCorpus & 0.325 & 0.260 & 0.278 & 0.277 & 0.328 & 0.308 & 0.342 & 0.308 & \textbf{0.346} \\
    \hline
    NQ & 0.329 & 0.467 & 0.446 & 0.486 & 0.498 & 0.495 & {0.568} & 0.518 & \textbf{0.570}   \\
    HotpotQA & 0.603 & 0.488 & 0.541 & 0.538 & 0.638 & 0.535 & 0.599 & 0.635 & \textbf{0.681}  \\
    FiQA-2018 & 0.236 & 0.252 & 0.259 & 0.259 & 0.329 & 0.349 & \textbf{0.467} & 0.316 & 0.345  \\
    \hline
    Signal-1M(RT) & 0.330 & 0.204 & 0.256 & 0.261 & 0.199 & 0.261 & \textbf{0.273} & 0.265 & 0.213  \\
    \hline
    TREC-NEWS & 0.398 & 0.362 & 0.358 & 0.376 & 0.428 & 0.337 & 0.346 & \textbf{0.428} & {0.427}  \\
    Robust04 & 0.408 & 0.351 & 0.365 & 0.349 & 0.476 & 0.437 & \textbf{0.506} & 0.447 & {0.479} \\
    \hline
    ArguAna & 0.315 & 0.265 & 0.389 & 0.298 & 0.446 & 0.511 & \textbf{0.540} & 0.433 & 0.474  \\
    Touche-2020 & \textbf{0.367} & 0.259 & 0.225 & 0.248 & 0.204 & 0.205 & 0.256 & 0.237 & 0.343  \\
    \hline
    CQADupStack & 0.299 & 0.282 & 0.290 & 0.347 & 0.345 & 0.357 & \textbf{0.399} & 0.317 & 0.320  \\
    Quora & 0.789 & 0.787 & 0.852 & 0.853 & 0.865 & 0.881 & \textbf{0.892} & 0.847 & 0.845  \\
    \hline
    DBPedia & 0.313 & 0.314 & 0.330 & 0.339 & {0.413} & 0.347 & 0.408 & 0.390 & \textbf{0.418} \\
    \hline
    SCIDOCS & 0.158 & 0.113 & 0.124 & 0.133 & \textbf{0.165} & 0.149 & 0.161 & 0.150 & 0.153  \\
    \hline
    FEVER & 0.753 & 0.682 & 0.641 & 0.691 & 0.758 & 0.660 & 0.740 & 0.774 & \textbf{0.800}  \\
    Climate-FEVER & 0.213 & 0.187 & 0.176 & 0.211 & 0.237 & 0.241 & \textbf{0.267} & 0.232 & 0.232  \\
    SciFact & 0.665 & 0.533 & 0.575 & 0.593 & 0.677 & 0.600 & 0.662 & 0.653 & \textbf{0.699}   \\
    \hhline{=|=|=|=|=|=|=|=|=|=}
    AVERAGE & 0.423 & 0.371 & 0.391 & 0.407 & 0.448 & 0.416 & 0.458 & 0.452 & \textbf{0.475} \\
    \ChangeRT{1pt}
    \end{tabular}
    \caption{Zero-shot retrieval performances on BEIR benchmark (measured by NDCG@10).} 
    \label{tab:3}
\end{table*}

\subsection{Main Results} 
The supervised evaluation results on MS MARCO and the zero-shot evaluation results on BEIR are reported in Table \ref{tab:2} and \ref{tab:3}, respectively. 

For supervised evaluations, we compare two groups baselines: (1) the models fine-tuned by hard negative samples (shown in the upper box), and (2) the models fine-tuned by knowledge distillation (shown in the lower box). As introduced, the knowledge distillation based fine-tuning are much more expensive, considering that it calls for the training and scoring of a teacher model. The two groups are corresponding to DupMAE (stage 2) and DupMAE (stage 3), which follow the same fine-tuning strategies. 

We may make the following observations from the supervised evaluations in Table \ref{tab:2}. Firstly, DupMAE gives rise to superior retrieval performances on MS MARCO. In particularly, DupMAE (stage 3) outperforms all baselines by notable advantages, and it further improves RetroMAE by almost \textbf{+1\%} absolute point, reaching a MRR@10 of \textbf{0.426}. Secondly, the retrieval quality may benefit substantially from the expensive knowledge distillation based fine-tuning, as the methods in the middle box are generally better than those in the upper box. However, we may also observe that with our enhanced pre-trained model, DupMAE (stage 2) is able to outperform the majority of the baseline methods with relatively low-cost fine-tuning operations, i.e., purely with hard negative samples. The two observations jointly reflect DupMAE's value to real-world applications: it will not only improve the best performance where neural retrievers may get, but also help to produce strong retrieval quality in a cost-effective way. 

For zero-shot settings, we compare DupMAE against the typical sparse retriever BM25 and other neural retrievers based on pre-trained language models. We report the retrieval performance on every single dataset, and measure the overall performance by taking the average of all 18 datasets. 

We may derive the following observations from the evaluation results in Table \ref{tab:3}. Firstly, DupMAE achieves remarkable performance on BEIR, reaching a NDCG@10 of \textbf{0.475} in total average of all 18 datasets. It outperforms its peer method RetroMAE on 13 out 18 datasets, and by \textbf{+2.3\%} absolute points in total average. Secondly, it can be observed that BM25 is a quite competitive baseline in reality, which notably outperforms general pre-trained models in terms of overall retrieval quality. Even with the massive-scale GTR-XXL, which uses as much as 4.8 billion model parameters and huge amounts of pre-training data, it still loses to BM25 on 8 out 18 datasets. However, with DupMAE, we may outperform BM25 on 15 out 18 datasets, and achieves \textbf{+5.2\%} absolute points improvement in total average. The above performances are impressive considering that DupMAE is merely based on a BERT-base scale encoder and uses much less pre-training data compared with other strong baselines, like Contriever and GTR. 

Given the analysis towards the main experiment results in Table \ref{tab:2} and \ref{tab:3}, we may draw the following conclusions in response to \textbf{RQ 1} and \textbf{2}: 
\begin{itemize}
    \item \textbf{Con 1}. DupMAE achieves notably higher retrieval quality over RetroMAE, indicating that the pre-trained model's semantic representation capability is substantially enhanced. 
    \item \textbf{Con 2}. DupMAE preserves remarkable performances throughout different situations, i.e., supervised and zero-shot evaluation, hard negative samples and knowledge distillation based fine-tuning, which reflects DupMAE's strong usability in practice. 
\end{itemize} 

\begin{table*}[t]
    \centering
    \scriptsize
    \begin{tabular}{C{0.3cm}|p{2.4cm}|C{1.6cm}|C{1.6cm}|C{1.6cm}|C{1.6cm}|C{1.6cm} }
    \ChangeRT{1pt} 
    & & 
    \multicolumn{5}{c}{\textbf{MS MARCO Dev}}  \\
    \hline
    & \textbf{Methods} & 
    \textbf{MRR@10} & \textbf{MRR@100} & \textbf{R@10} & \textbf{R@100} & \textbf{R@1000}  \\
    \hline
    \multirow{4}{*}{1.} 
    & RetroMAE & 0.3928 & 0.4032 & 0.6749& 0.9178 & 0.9849 \\
    & CLS decoding only & 0.4008 & 0.4099 & 0.6906 & 0.9229 & 0.9840   \\
    & OT decoding only & 0.4002 & 0.4092 & 0.6890 & 0.9213 & 0.9831   \\
    & CLS and OT decoding & \textbf{0.4102} & \textbf{0.4202} & \textbf{0.7049} & \textbf{0.9280} & \textbf{0.9874}  \\ 
    \hhline{=|=|=|=|=|=|=}
    \multirow{4}{*}{2.} 
    & CLS:768 & 0.3941 & 0.4040 & 0.6865 & 0.9174 & 0.9871  \\ 
    & OT:768 & 0.4019 & 0.4114 & 0.6934 & 0.9095 & 0.9814  \\ 
    & CLS:384, OT:384 & \textbf{0.4102} & \textbf{0.4202} & \textbf{0.7049} & 0.9280 & 0.9874   \\ 
    & CLS:384, OT:260 & 0.4071 & 0.4171 & 0.7037 & \textbf{0.9293} & \textbf{0.9882}  \\ 
    \ChangeRT{1pt}
    \end{tabular}
    \vspace{-5pt}
    \caption{Ablation studies: 1. impact from pre-training, 2. impact from semantic representations.} 
    \vspace{-10pt}
    \label{tab:distill}
\end{table*}


\subsection{Ablation Studies} 
After the verification of DupMAE's overall effectiveness, it remains to figure out which factors contribute to its improvements over the existing methods. To this end, we perform ablation studies for DupMAE, whose results are shown in Table \ref{tab:distill}. We use MS MARCO dataset for our exploration, and leverage hard negative samples to fine-tune the pre-trained models.  

We conduct two sets of experiments for our ablation studies. First of all, we explore \textbf{the impact from pre-training}, whose results are shown in the upper part of Table \ref{tab:distill}. Remember that DupMAE includes two decoding tasks as discussed in Section \ref{sec:method-dup}: CLS decoding and OT decoding, we make evaluations for three alternative forms accordingly. (1) CLS decoding only, which only preserves the pre-training task for the [CLS] embedding; (2) OT decoding only, which only performs the pre-training task for the OT embeddings; (3) CLS and OT decoding, which is exactly the pre-training method used by DupMAE. We also introduce RetroMAE for comparison. Note that RetroMAE and ``CLS decoding only'' share the same pre-training task; however, their semantic representations are generated in different ways. 

We may derive the following observations from the experiment results. Firstly, the joint utilization of the two pre-training tasks leads to the optimal retrieval quality, where the MRR@10 grows beyond ``CLS only'' and ``OT only'' by almost +1\% absolute point. As a result, the effectiveness of both pre-training tasks and their joint effect can be verified. Secondly, RetroMAE's performance is inferior to other methods, especially ``CLS pre-train only'' which share the pre-training task with it. Such an observation can be attribute to the difference on semantic representation: DupMAE relies on the contextualized embeddings from both [CLS] and ordinary tokens, while RetroMAE only leverages the [CLS] token's embedding. 

We further explore \textbf{the impact from semantic representation} with the lower part of Table \ref{tab:distill}). As introduced in Section \ref{sec:method-dup}, DupMAE's default semantic representation (dim-768) consists of two parts: half of its 768 float values come from the linear projection of [CLS] embedding, while the other half come from the sparsification of OT embeddings (denoted as ``CLS:384, OT:384''). In this place, we consider two variational formulations: (1) ``CLS:768'', which directly uses the [CLS] embedding, and (2) ``OT:768'', where the OT embeddings are sparsified to 768 float values for the representation of the input. According to the experiment results, the performance of ``OT:768'' is slightly better than ``CLS:768''. At the same time, the default DupMAE, ``CLS:384, OT:384'', results in a better performance than both variational formulations. The above observations indicate that the contextualized embeddings from [CLS] and ordinary tokens may provide complementary information about the input data. Therefore, the joint utilization of both types of embeddings is able to generate more a powerful semantic representation. 

Note that ``CLS:384, OT:384'' preserves the same similarity computation cost as ``CLS:768'' (the inner product between query and document representations); however, it's storage cost is slightly higher than ``CLS:768'', as extra space is resulted from the indexes of OT embeddings' sparsification results. (Each index will take less than 15 bits knowing that the vocabulary space is 30522.) In this place, we introduce another variation ``CLS:384,OT:260'' by further reducing the dimension of OT embeddings, which leads to the same storage cost as ``CLS:768''. It can be observed that ``CLS:384,OT:260'' outperforms the previous two variations, and preserves a similar performance as ``CLS:384, OT:384''. 

Given the analysis towards the ablation studies in Table \ref{tab:distill}, we may come to the following conclusions in response to \textbf{RQ 3} and \textbf{4}: 
\begin{itemize}
    \item \textbf{Con 3}. The [CLS] and OT embeddings may provide complementary information to each other; the aggregation of them helps to generate stronger semantic representations. 
    \item \textbf{Con 4}. Both tasks: [CLS] and OT decoding, contribute to DupMAE; the joint utilization of both tasks lead to the optimal performance.   
\end{itemize}




\section{Conclusion} 
This paper presents DupMAE, a new retrieval-oriented pre-training method where the semantic representation capacities can be jointly enhanced for all contextualized embeddings of the language models. It leverages RetroMAE's decoding task for [CLS]'s embedding and introduces a BoW-based decoding task for OT embeddings. The training losses from the two tasks are added up for a unified encoder. The two types of embeddings, after dimension reduction, are aggregated for the joint semantic representation. The effectiveness of our proposed method is empirically verified, as remarkable retrieval performances are achieved on MS MARCO and BEIR throughout different situations. 


\newpage
\bibliographystyle{acl_natbib}
\bibliography{main}

\begin{thebibliography}{36}
\expandafter\ifx\csname natexlab\endcsname\relax\def\natexlab#1{#1}\fi

\bibitem[{Chang et~al.(2020)Chang, Yu, Chang, Yang, and Kumar}]{chang2020pre}
Wei-Cheng Chang, Felix~X Yu, Yin-Wen Chang, Yiming Yang, and Sanjiv Kumar.
  2020.
\newblock Pre-training tasks for embedding-based large-scale retrieval.
\newblock \emph{arXiv preprint arXiv:2002.03932}.

\bibitem[{Craswell et~al.(2020)Craswell, Mitra, Yilmaz, Campos, and
  Voorhees}]{craswell2020overview}
Nick Craswell, Bhaskar Mitra, Emine Yilmaz, Daniel Campos, and Ellen~M
  Voorhees. 2020.
\newblock Overview of the trec 2019 deep learning track.
\newblock \emph{arXiv preprint arXiv:2003.07820}.

\bibitem[{Devlin et~al.(2019)Devlin, Chang, Lee, and
  Toutanova}]{Devlin2019BERT}
Jacob Devlin, Ming{-}Wei Chang, Kenton Lee, and Kristina Toutanova. 2019.
\newblock {BERT:} pre-training of deep bidirectional transformers for language
  understanding.
\newblock In \emph{Proceedings of the 2019 Conference of the North American
  Chapter of the Association for Computational Linguistics: Human Language
  Technologies,}, pages 4171--4186. Association for Computational Linguistics.

\bibitem[{Formal et~al.(2021)Formal, Lassance, Piwowarski, and
  Clinchant}]{formal2021splade}
Thibault Formal, Carlos Lassance, Benjamin Piwowarski, and St{\'e}phane
  Clinchant. 2021.
\newblock Splade v2: Sparse lexical and expansion model for information
  retrieval.
\newblock \emph{arXiv preprint arXiv:2109.10086}.

\bibitem[{Gao and Callan(2021)}]{gao2021condenser}
Luyu Gao and Jamie Callan. 2021.
\newblock Condenser: a pre-training architecture for dense retrieval.
\newblock In \emph{Proceedings of the 2021 Conference on Empirical Methods in
  Natural Language Processing}, pages 981--993.

\bibitem[{Gao and Callan(2022)}]{gao2021unsupervised}
Luyu Gao and Jamie Callan. 2022.
\newblock Unsupervised corpus aware language model pre-training for dense
  passage retrieval.
\newblock In \emph{Proceedings of the 60th Annual Meeting of the Association
  for Computational Linguistics (Volume 1: Long Papers)}, pages 2843--2853,
  Dublin, Ireland.

\bibitem[{Guu et~al.(2020)Guu, Lee, Tung, Pasupat, and Chang}]{guu2020realm}
Kelvin Guu, Kenton Lee, Zora Tung, Panupong Pasupat, and Ming-Wei Chang. 2020.
\newblock Realm: Retrieval-augmented language model pre-training.
\newblock \emph{arXiv preprint arXiv:2002.08909}.

\bibitem[{Hofst{\"a}tter et~al.(2021)Hofst{\"a}tter, Lin, Yang, Lin, and
  Hanbury}]{hofstatter2021efficiently}
Sebastian Hofst{\"a}tter, Sheng-Chieh Lin, Jheng-Hong Yang, Jimmy Lin, and
  Allan Hanbury. 2021.
\newblock Efficiently teaching an effective dense retriever with balanced topic
  aware sampling.
\newblock In \emph{Proceedings of the 44th International ACM SIGIR Conference
  on Research and Development in Information Retrieval}, pages 113--122.

\bibitem[{Huang et~al.(2013)Huang, He, Gao, Deng, Acero, and
  Heck}]{huang2013learning}
Po-Sen Huang, Xiaodong He, Jianfeng Gao, Li~Deng, Alex Acero, and Larry Heck.
  2013.
\newblock Learning deep structured semantic models for web search using
  clickthrough data.
\newblock In \emph{Proceedings of the 22nd ACM international conference on
  Information \& Knowledge Management}, pages 2333--2338.

\bibitem[{Humeau et~al.(2019)Humeau, Shuster, Lachaux, and
  Weston}]{humeau2019poly}
Samuel Humeau, Kurt Shuster, Marie-Anne Lachaux, and Jason Weston. 2019.
\newblock Poly-encoders: Transformer architectures and pre-training strategies
  for fast and accurate multi-sentence scoring.
\newblock \emph{arXiv preprint arXiv:1905.01969}.

\bibitem[{Izacard et~al.(2021)Izacard, Caron, Hosseini, Riedel, Bojanowski,
  Joulin, and Grave}]{izacard2021towards}
Gautier Izacard, Mathilde Caron, Lucas Hosseini, Sebastian Riedel, Piotr
  Bojanowski, Armand Joulin, and Edouard Grave. 2021.
\newblock Towards unsupervised dense information retrieval with contrastive
  learning.
\newblock \emph{arXiv preprint arXiv:2112.09118}.

\bibitem[{Izacard et~al.(2022)Izacard, Lewis, Lomeli, Hosseini, Petroni,
  Schick, Dwivedi-Yu, Joulin, Riedel, and Grave}]{izacard2022few}
Gautier Izacard, Patrick Lewis, Maria Lomeli, Lucas Hosseini, Fabio Petroni,
  Timo Schick, Jane Dwivedi-Yu, Armand Joulin, Sebastian Riedel, and Edouard
  Grave. 2022.
\newblock Few-shot learning with retrieval augmented language models.
\newblock \emph{arXiv preprint arXiv:2208.03299}.

\bibitem[{Karpukhin et~al.(2020)Karpukhin, Oguz, Min, Lewis, Wu, Edunov, Chen,
  and Yih}]{karpukhin2020dense}
Vladimir Karpukhin, Barlas Oguz, Sewon Min, Patrick Lewis, Ledell Wu, Sergey
  Edunov, Danqi Chen, and Wen-tau Yih. 2020.
\newblock Dense passage retrieval for open-domain question answering.
\newblock In \emph{Proceedings of the 2020 Conference on Empirical Methods in
  Natural Language Processing}, pages 6769--6781.

\bibitem[{Komeili et~al.(2021)Komeili, Shuster, and
  Weston}]{komeili2021internet}
Mojtaba Komeili, Kurt Shuster, and Jason Weston. 2021.
\newblock Internet-augmented dialogue generation.
\newblock \emph{arXiv preprint arXiv:2107.07566}.

\bibitem[{Lin et~al.(2022)Lin, Li, and Lin}]{lin2022aggretriever}
Sheng-Chieh Lin, Minghan Li, and Jimmy Lin. 2022.
\newblock Aggretriever: A simple approach to aggregate textual representation
  for robust dense passage retrieval.
\newblock \emph{arXiv preprint arXiv:2208.00511}.

\bibitem[{Liu et~al.(2019)Liu, Ott, Goyal, Du, Joshi, Chen, Levy, Lewis,
  Zettlemoyer, and Stoyanov}]{Liu2019Roberta}
Yinhan Liu, Myle Ott, Naman Goyal, Jingfei Du, Mandar Joshi, Danqi Chen, Omer
  Levy, Mike Lewis, Luke Zettlemoyer, and Veselin Stoyanov. 2019.
\newblock Roberta: {A} robustly optimized {BERT} pretraining approach.
\newblock \emph{CoRR}, abs/1907.11692.

\bibitem[{Lu et~al.(2021)Lu, He, Xiong, Ke, Malik, Dou, Bennett, Liu, and
  Overwijk}]{lu2021less}
Shuqi Lu, Di~He, Chenyan Xiong, Guolin Ke, Waleed Malik, Zhicheng Dou, Paul
  Bennett, Tie-Yan Liu, and Arnold Overwijk. 2021.
\newblock Less is more: Pretrain a strong {S}iamese encoder for dense text
  retrieval using a weak decoder.
\newblock In \emph{Proceedings of the 2021 Conference on Empirical Methods in
  Natural Language Processing}, pages 2780--2791.

\bibitem[{Lu et~al.(2022)Lu, Liu, Liu, Shi, Huang, Sun, Tian, Wu, Wang, Yin
  et~al.}]{lu2022ernie}
Yuxiang Lu, Yiding Liu, Jiaxiang Liu, Yunsheng Shi, Zhengjie Huang, Shikun
  Feng~Yu Sun, Hao Tian, Hua Wu, Shuaiqiang Wang, Dawei Yin, et~al. 2022.
\newblock Ernie-search: Bridging cross-encoder with dual-encoder via self
  on-the-fly distillation for dense passage retrieval.
\newblock \emph{arXiv preprint arXiv:2205.09153}.

\bibitem[{Luan et~al.(2021)Luan, Eisenstein, Toutanova, and
  Collins}]{luan2021sparse}
Yi~Luan, Jacob Eisenstein, Kristina Toutanova, and Michael Collins. 2021.
\newblock Sparse, dense, and attentional representations for text retrieval.
\newblock \emph{Transactions of the Association for Computational Linguistics},
  9:329--345.

\bibitem[{Nguyen et~al.(2016)Nguyen, Rosenberg, Song, Gao, Tiwary, Majumder,
  and Deng}]{nguyen2016ms}
Tri Nguyen, Mir Rosenberg, Xia Song, Jianfeng Gao, Saurabh Tiwary, Rangan
  Majumder, and Li~Deng. 2016.
\newblock Ms marco: A human generated machine reading comprehension dataset.
\newblock In \emph{CoCo@ NIPS}.

\bibitem[{Ni et~al.(2021)Ni, Qu, Lu, Dai, {\'A}brego, Ma, Zhao, Luan, Hall,
  Chang et~al.}]{ni2021large}
Jianmo Ni, Chen Qu, Jing Lu, Zhuyun Dai, Gustavo~Hern{\'a}ndez {\'A}brego,
  Ji~Ma, Vincent~Y Zhao, Yi~Luan, Keith~B Hall, Ming-Wei Chang, et~al. 2021.
\newblock Large dual encoders are generalizable retrievers.
\newblock \emph{arXiv preprint arXiv:2112.07899}.

\bibitem[{Qu et~al.(2020)Qu, Ding, Liu, Liu, Ren, Zhao, Dong, Wu, and
  Wang}]{qu2020rocketqa}
Yingqi Qu, Yuchen Ding, Jing Liu, Kai Liu, Ruiyang Ren, Wayne~Xin Zhao, Daxiang
  Dong, Hua Wu, and Haifeng Wang. 2020.
\newblock Rocketqa: An optimized training approach to dense passage retrieval
  for open-domain question answering.
\newblock \emph{arXiv preprint arXiv:2010.08191}.

\bibitem[{Raffel et~al.(2019)Raffel, Shazeer, Roberts, Lee, Narang, Matena,
  Zhou, Li, and Liu}]{raffel2019exploring}
Colin Raffel, Noam Shazeer, Adam Roberts, Katherine Lee, Sharan Narang, Michael
  Matena, Yanqi Zhou, Wei Li, and Peter~J Liu. 2019.
\newblock Exploring the limits of transfer learning with a unified text-to-text
  transformer.
\newblock \emph{arXiv preprint arXiv:1910.10683}.

\bibitem[{Ren et~al.(2021)Ren, Qu, Liu, Zhao, She, Wu, Wang, and
  Wen}]{ren2021rocketqav2}
Ruiyang Ren, Yingqi Qu, Jing Liu, Wayne~Xin Zhao, Qiaoqiao She, Hua Wu, Haifeng
  Wang, and Ji-Rong Wen. 2021.
\newblock Rocketqav2: A joint training method for dense passage retrieval and
  passage re-ranking.
\newblock \emph{arXiv preprint arXiv:2110.07367}.

\bibitem[{Santhanam et~al.(2021)Santhanam, Khattab, Saad-Falcon, Potts, and
  Zaharia}]{santhanam2021colbertv2}
Keshav Santhanam, Omar Khattab, Jon Saad-Falcon, Christopher Potts, and Matei
  Zaharia. 2021.
\newblock Colbertv2: Effective and efficient retrieval via lightweight late
  interaction.
\newblock \emph{arXiv preprint arXiv:2112.01488}.

\bibitem[{Thakur et~al.(2021)Thakur, Reimers, R{\"u}ckl{\'e}, Srivastava, and
  Gurevych}]{thakur2021beir}
Nandan Thakur, Nils Reimers, Andreas R{\"u}ckl{\'e}, Abhishek Srivastava, and
  Iryna Gurevych. 2021.
\newblock Beir: A heterogenous benchmark for zero-shot evaluation of
  information retrieval models.
\newblock \emph{arXiv preprint arXiv:2104.08663}.

\bibitem[{Wang et~al.(2021)Wang, Reimers, and Gurevych}]{wang2021tsdae}
Kexin Wang, Nils Reimers, and Iryna Gurevych. 2021.
\newblock Tsdae: Using transformer-based sequential denoising auto-encoder for
  unsupervised sentence embedding learning.
\newblock \emph{arXiv preprint arXiv:2104.06979}.

\bibitem[{Wang et~al.(2022)Wang, Yang, Huang, Jiao, Yang, Jiang, Majumder, and
  Wei}]{wang2022simlm}
Liang Wang, Nan Yang, Xiaolong Huang, Binxing Jiao, Linjun Yang, Daxin Jiang,
  Rangan Majumder, and Furu Wei. 2022.
\newblock Simlm: Pre-training with representation bottleneck for dense passage
  retrieval.
\newblock \emph{arXiv preprint arXiv:2207.02578}.

\bibitem[{Xiao and Liu(2022)}]{retromae-code}
Shitao Xiao and Zheng Liu. 2022.
\newblock \href {https://github.com/staoxiao/RetroMAE} {Retromae repo: the
  codebase for retrieval-oriented language models;
  https://github.com/staoxiao/retromae}.

\bibitem[{Xiao et~al.(2022{\natexlab{a}})Xiao, Liu, Shao, and
  Cao}]{liu2022retromae}
Shitao Xiao, Zheng Liu, Yingxia Shao, and Zhao Cao. 2022{\natexlab{a}}.
\newblock Retromae: Pre-training retrieval-oriented language models via masked
  auto-encoder.
\newblock \emph{arXiv preprint arXiv:2205.12035}.

\bibitem[{Xiao et~al.(2022{\natexlab{b}})Xiao, Liu, Shao, Di, Middha, Wu, and
  Xie}]{xiao2022training}
Shitao Xiao, Zheng Liu, Yingxia Shao, Tao Di, Bhuvan Middha, Fangzhao Wu, and
  Xing Xie. 2022{\natexlab{b}}.
\newblock Training large-scale news recommenders with pretrained language
  models in the loop.
\newblock In \emph{Proceedings of the 28th ACM SIGKDD Conference on Knowledge
  Discovery and Data Mining}, pages 4215--4225.

\bibitem[{Xiong et~al.(2020)Xiong, Xiong, Li, Tang, Liu, Bennett, Ahmed, and
  Overwijk}]{xiong2020approximate}
Lee Xiong, Chenyan Xiong, Ye~Li, Kwok-Fung Tang, Jialin Liu, Paul Bennett,
  Junaid Ahmed, and Arnold Overwijk. 2020.
\newblock Approximate nearest neighbor negative contrastive learning for dense
  text retrieval.
\newblock \emph{arXiv preprint arXiv:2007.00808}.

\bibitem[{Zhan et~al.(2021)Zhan, Mao, Liu, Guo, Zhang, and
  Ma}]{zhan2021optimizing}
Jingtao Zhan, Jiaxin Mao, Yiqun Liu, Jiafeng Guo, Min Zhang, and Shaoping Ma.
  2021.
\newblock Optimizing dense retrieval model training with hard negatives.
\newblock In \emph{Proceedings of the 44th International ACM SIGIR Conference
  on Research and Development in Information Retrieval}, pages 1503--1512.

\bibitem[{Zhang et~al.(2021)Zhang, Gong, Shen, Lv, Duan, and
  Chen}]{zhang2021adversarial}
Hang Zhang, Yeyun Gong, Yelong Shen, Jiancheng Lv, Nan Duan, and Weizhu Chen.
  2021.
\newblock Adversarial retriever-ranker for dense text retrieval.
\newblock \emph{arXiv preprint arXiv:2110.03611}.

\bibitem[{Zhang et~al.(2022)Zhang, Liu, Han, Xiao, Zheng, Shao, Sun, Zhu,
  Srinivasan, Deng et~al.}]{zhang2022uni}
Jianjin Zhang, Zheng Liu, Weihao Han, Shitao Xiao, Ruicheng Zheng, Yingxia
  Shao, Hao Sun, Hanqing Zhu, Premkumar Srinivasan, Weiwei Deng, et~al. 2022.
\newblock Uni-retriever: Towards learning the unified embedding based retriever
  in bing sponsored search.
\newblock In \emph{Proceedings of the 28th ACM SIGKDD Conference on Knowledge
  Discovery and Data Mining}, pages 4493--4501.

\bibitem[{Zhou et~al.(2022)Zhou, Gong, Liu, Zhao, Shen, Dong, Lu, Majumder,
  Wen, Duan et~al.}]{zhou2022simans}
Kun Zhou, Yeyun Gong, Xiao Liu, Wayne~Xin Zhao, Yelong Shen, Anlei Dong,
  Jingwen Lu, Rangan Majumder, Ji-Rong Wen, Nan Duan, et~al. 2022.
\newblock Simans: Simple ambiguous negatives sampling for dense text retrieval.
\newblock \emph{arXiv preprint arXiv:2210.11773}.

\end{thebibliography}

\end{document}